\pdfoutput=1
\documentclass[11pt]{article}
\usepackage[]{EMNLP2023}

\usepackage{times}
\usepackage{latexsym}

\usepackage[utf8]{inputenc}
\usepackage[T1]{fontenc}

\usepackage{microtype}

\usepackage{inconsolata}
\usepackage{graphicx}
\usepackage{booktabs}
\usepackage{algorithm}
\usepackage{algpseudocode}
\usepackage{multirow} 
\usepackage{makecell}
\usepackage{amsmath,amssymb,mathabx,mathtools,amsthm,nicefrac}
\usepackage[capitalise,noabbrev,nameinlink]{cleveref}
\usepackage{enumitem}
\usepackage{setspace}
\usepackage{xcolor}
\title{Learning from Mistakes: Iterative Prompt Relabeling\\ for Text-to-Image Diffusion Model Training} 

\author{
    \normalfont
    \textbf{Xinyan Chen}\textsuperscript{1,2,$\star$} \quad 
    \textbf{Jiaxin Ge}\textsuperscript{1,$\star$} \quad 
    \textbf{Tianjun Zhang}\textsuperscript{3,$\star$} \\ 
    \textbf{Jiaming Liu}\textsuperscript{1} \quad 
    \textbf{Shanghang Zhang}\textsuperscript{1,$\dagger,\ddagger$}
    \vspace{0.3em}\\
    \normalfont
    $^\star$Equal contributions \quad $^\dagger$Corresponding author\\
    $^1$State Key Laboratory of Multimedia Information Processing,\\ School of Computer Science, Peking University\\  $^2$ University of Science and Technology of China \quad
    $^3$UC Berkeley\\
    $^\ddagger$\texttt{shanghang@pku.edu.cn}
}

\newcommand{\myparagraph}[1]{\paragraph{#1}}

\begin{document}
\maketitle
\begin{abstract}
Diffusion models have shown impressive performance in many domains. 
However, the model's capability to follow natural language instructions (e.g., spatial relationships between objects, generating complex scenes) is still unsatisfactory. 
In this work, we propose \textbf{I}terative \textbf{P}rompt \textbf{R}elabeling (IPR), a novel algorithm that aligns images to text through iterative image sampling and prompt relabeling \textit{with feedback}. 
IPR first samples a batch of images conditioned on the text, then relabels the text prompts of unmatched text-image pairs with classifier feedback. 
We conduct thorough experiments on SDv2 and SDXL, testing their capability to follow instructions on spatial relations. With IPR, we improved up to 15.22\% (absolute improvement) on the challenging spatial relation VISOR benchmark, demonstrating superior performance compared to previous RL methods. Our code is publicly available at \href{https://github.com/xinyan-cxy/IPR-RLDF}{https://github.com/xinyan-cxy/IPR-RLDF}.
\end{abstract}

\section{Introduction}\label{sec:intro}
\setstretch{0.98}
Recent advancements in the field of image generation have been notably driven by diffusion models, especially in the area of text-to-image conversion \citep{ho2020denoising}. 
However, a significant challenge arises when these models are tasked with interpreting and executing complex instructions, particularly those involving spatial relationships \citep{gal2022image, saharia2022photorealistic}. 
One simple example is that a prompt like ``a dog left to a car'' often results in images where the spatial relationship ``left to'' is not accurately depicted. 
This limitation underscores a crucial gap in the current models' ability to understand and render intricate spatial relationships.

\begin{figure*}[ht]
    \centering
    \includegraphics[width=0.90\linewidth]{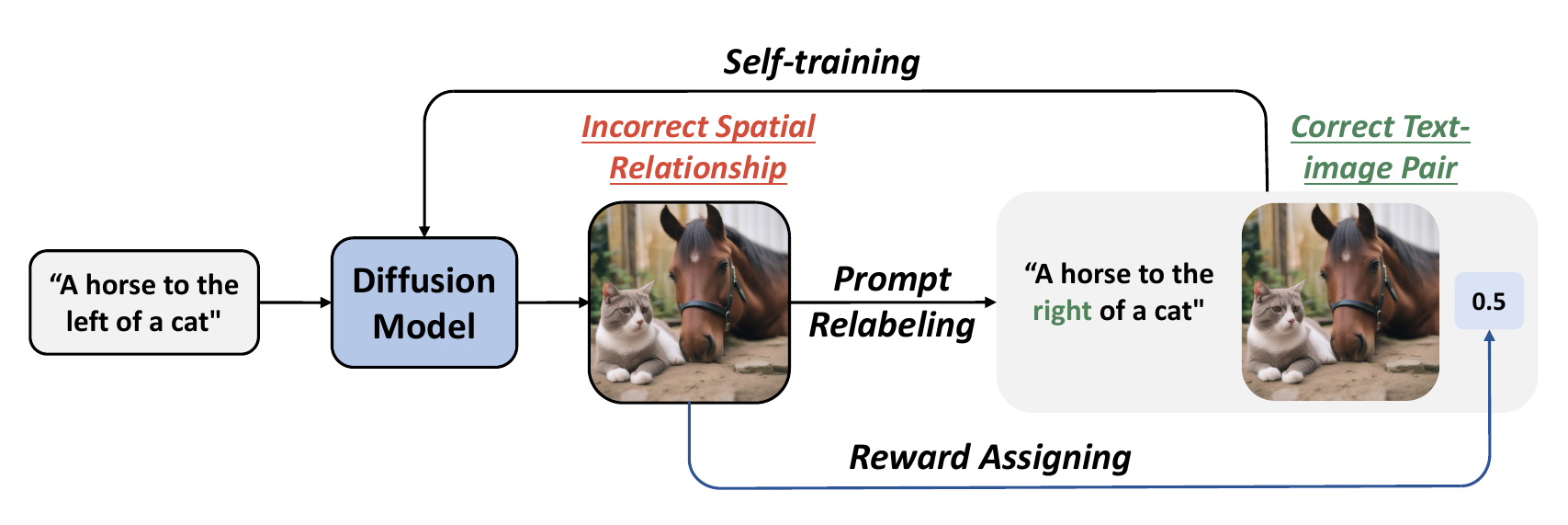}
    \caption{\textbf{A high-level overview of our approach.} We enhance the alignment of images with text through an iterative process of image sampling and prompt relabeling.}
    \label{fig:high_level_motivation}
\end{figure*}


In response to these challenges, we investigate how models can self-improve without using extensive training data. Some studies have found that self-supervised learning is useful for training diffusion models \citep{hu2023selfguided}. However, these methods still lack an effective self-correction mechanism, making the self-generated dataset either noisy or being trimmed off too much. 
Recent advances in large language models (LLM) have found that learning from mistakes and receiving language feedback can effectively enhance the model's reasoning capacity \citep{an2024learning}. Inspired by this, in our research, we study whether we can effectively use the originally incorrect text-image pairs to train the diffusion model.

In particular, we propose \textbf{I}terative \textbf{P}rompt \textbf{R}elabeling (IPR), a novel technique designed to enhance the alignment of images with text through an iterative process of image sampling and prompt relabeling. Our approach uses rich language feedback for all the images and a simple reward design. 
A high-level overview of our approach is shown in \Cref{fig:high_level_motivation}.
We begin by establishing a reward function based on an external detection model that automatically classifies the correctness of an image and assigns rewards based on the detection results. 
This enables a straightforward reward design and neglects the complicated training of reward models. 
Then, we use prompt relabeling to relabel the input prompt of the mismatched image-text pairs based on the results of the detection model. 
This allows the models to use the mis-generated images with the correct version. 
Finally, we adopt iterative training that continuously trains the model with its self-generated images. Iterative training allows dynamically scaling up the dataset and receiving additional feedback through multiple training rounds, which can further enrich the feedback and progressively refine the model's performance.

We demonstrate the efficacy of IPR across SDv2 \citep{rombach2022highresolution} and SDXL \citep{podell2023sdxl}, trained with and w/o LoRA, and we test the performance on the challenging spatial relation task, where we observe a substantial improvement of up to 15.22\% (absolute improvement) on the VISOR \citep{gokhale2023benchmarking} benchmark. This performance underlines the potential of IPR in pushing the boundaries of text-to-image generation models, especially in terms of understanding and rendering complex spatial relationships.
\vspace{-1mm}
\section{Related Work}
\vspace{-1mm}
Text-to-image models have facilitated the generation of high-resolution, multi-styled images \citep{saharia2022photorealistic, ramesh2021zeroshot, zhang2023controllable}. However, training diffusion models from scratch demands a substantial amount of data and time. Therefore, various fine-tuning strategies have been explored. These approaches include associating a unique identifier with a particular subject \citep{ruiz2023dreambooth}, introducing new embeddings to represent user-provided concepts \citep{gal2022image}, adapting compositional generation \citep{liu2023compositional}, and implementing Low-Rank Adaptation \citep{hu2021lora}. Our work utilizes LoRA to fine-tune the diffusion model and follow the reinforcement learning line of work to rescale loss for effective training.

Recent research has demonstrated the efficacy of prompt relabeling techniques. TEMPERA \citep{zhang2022tempera} offers interpretable prompts tailored to various queries through the creation of an innovative action space, enabling flexible adjustments to the initial prompts. Furthermore, the Hindsight Instruction Relabeling (HIR) \citep{zhang2023wisdom} approach conceptualizes the instruction alignment problem as a goal-reaching problem within the context of decision-making. It entails the conversion of feedback into instructions by re-labeling the original instructions. Our approach represents a significant advancement as it employs prompt relabeling techniques with diffusion models for the first time.

\section{Method}
\begin{figure*}[htbp]
    \centering
    \includegraphics[width=0.95\linewidth]{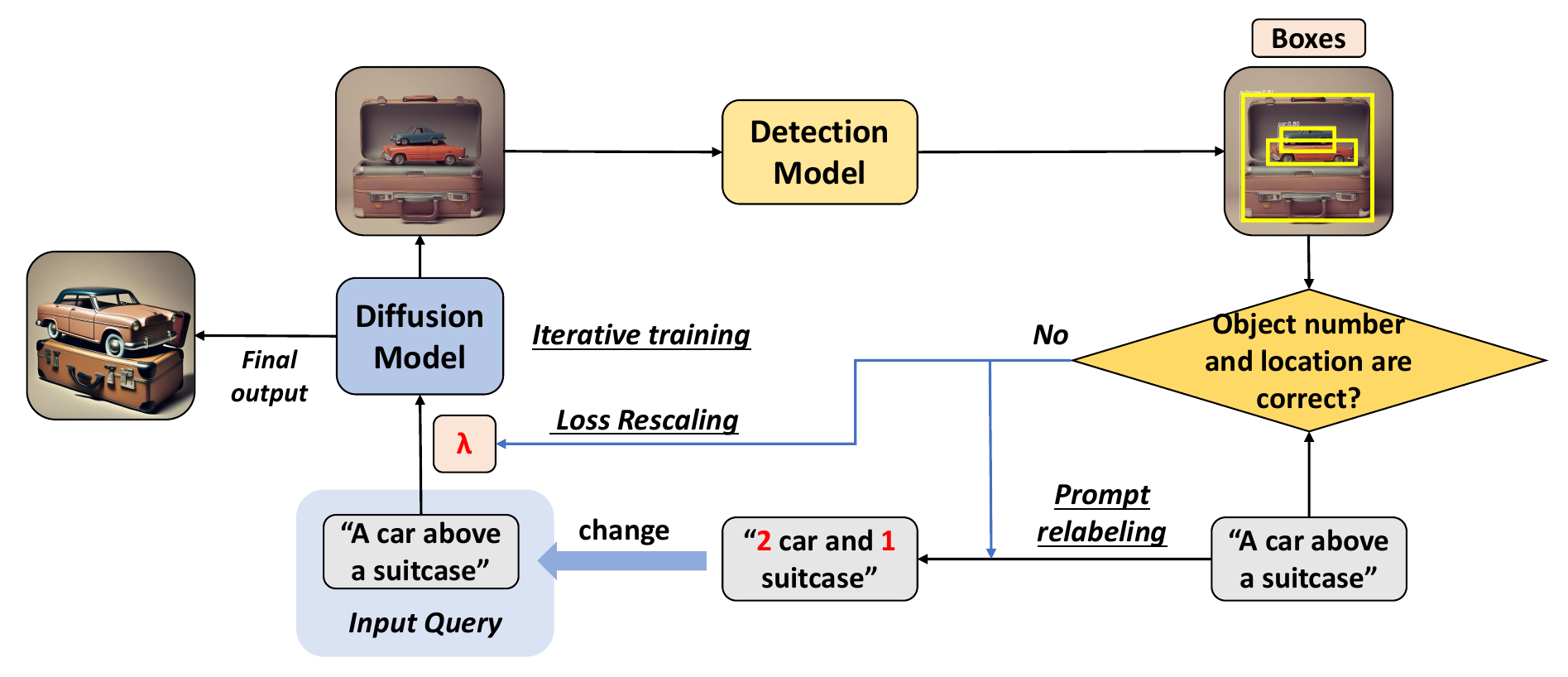}
    \caption{\textbf{The general pipeline of IPR.} Our approach adopts four different stages: (1) diffusion model sampling, (2) reward-based loss rescaling, (3) prompt relabeling, and (4) iterative training.}
    \label{fig:Pipeline}
\end{figure*}

    
    
    
\subsection{Method Overview} 
Our approach adopts four different stages: diffusion model sampling, reward-based loss rescaling, prompt relabeling, and iterative training. 
Our pipeline is demonstrated in \Cref{fig:Pipeline}. First, IPR samples a batch of images from the diffusion model using the text prompts. 
Then, IPR uses a detection model to check the correctness of the text-image pair and relabels the original text prompt to attain a correct text-image pair.
Next, IRP assigns a reward to the text-image pair based on the correctness of the original text-image pair.
Finally, IRP trains the diffusion model on the new relabeled text-image pairs in an iterative manner. 
We will introduce the details of each stage in the following sections.
\begin{table}[t]
    \small
    \centering
    \caption{Combined results of spatial accuracy and CLIP score on four RLDF settings. (1) Direct: The original diffusion models. (2) RLDF: Only applying RLDF on diffusion models. (3) IPR-RLDF: Our method (with RLDF). The results show that IPR outperforms other methods in generating images with correct spatial relationships.}
    \label{tab:combined_results}
    \resizebox{.95\linewidth}{!}{%
    \begin{tabular}{@{}llcccc@{}}
        \toprule
        Score Type & Method & SDv2(1) & \makecell{SDv2\\(LoRA)} & \makecell{SDXL\\(LoRA)} & SDv2(2)\\
        \midrule
        \multirow{3}{*}{\makecell{Spatial\\Accuracy (\%)}}
        & Direct & 18.75 & 18.75 & 27.00 &  17.00 \\
        & RLDF & 21.50 &22.00 & 29.75 & 22.44 \\
        & \textbf{IPR-RLDF} & \textbf{28.50} & \textbf{25.25} & \textbf{31.25} & \textbf{32.22} \\
        \midrule
        \multirow{3}{*}{CLIP Score} 
        & Direct & 25.75 & 25.75 & 27.41 & - \\
        & RLDF & \textbf{26.67} & 26.09 & 28.68 & - \\
        & \textbf{IPR-RLDF} & 25.87 & \textbf{26.15} & \textbf{28.74} & \textbf{-} \\
        \bottomrule
    \end{tabular}
    }
\end{table} 

\subsection{Prompt Relabeling}
The sampled images from the diffusion model consist of the ones that are aligned with the text prompt and the ones that are not. 
We then relabel the inconsistent image-text pairs to ensure that the textual description accurately reflects the content of the generated image. 
The algorithm consists of the following:
\begin{enumerate}[leftmargin=*,noitemsep]
    \item Detect objects in the generated image using a detection model, yielding bounding boxes (bbox) for each object.
    \item Compare the number of detected objects with the object count specified in the original prompt.
    \item If the object count matches, analyze the bounding box centers to determine the actual spatial relationship between objects. Modify the original prompt to accurately describe this relationship.
    \item If the object count does not match, revise the prompt to reflect the actual number and type of objects in the image (e.g., ``2 cats and 1 dog'').
\end{enumerate}

\subsection{Detection-Based Loss Rescaling}


In our method, when attaining the relabeled text-image pairs, we assign a reward-rescaling factor to each pair. This reward is used to rescale the loss when training the diffusion model.
The loss function is rescaled based on the following $
L_{\text{rescaled}} = \mathbb{E}_{x_i, c_i} \left[ \lambda_{x_i, c_i} L_{\theta}(x_i, c_i)  \right]$, where \( L_\theta \) is the standard diffusion model loss as defined in \citep{ho2020denoising}. The modulatory factor \( \lambda \) is defined as: $1$, when the model correctly generates the text-image pair; $0.5$, when the text-image pair is incorrect and needs relabeling.

The rescaling factor $\lambda$ balances the model's capability to follow text prompts with the model's generated image distribution not too far away from the original one. A larger $\lambda$ pushes the model to learn from the mis-generated text-image pairs while a smaller $\lambda$ keeps the model closer to its original distribution. After the rescaling weight is assigned to each text-image pair, we obtain a dataset of the model's self-generated text-image pairs. We use this dataset to train the diffusion model.
\subsection{Iterative Training}
The self-generated dataset is often not large enough. So we train the diffusion model by repeating this process iteratively. Specifically, after training the model for an iteration, we use the updated model to repeat the previous process and get a new dataset that contains text-image pairs and their corresponding rescaling weight. Then we use the new dataset to train on the new model to get the updated model for the next iteration. The training procedure is formulated as follows:
$\theta_{\text{new}} = \text{Train}(\theta_{\text{old}}, \{(x_{0}^{(i)}, c_{\text{new}}^{(i)}, L_{\text{rescaled}}^{(i)})\}_{i=1}^{N})$
where \( N \) denotes the number of samples in each iteration. This iterative process leads to continual improvement in the model's ability to produce spatially coherent images aligned with textual descriptions.

\begin{table}[t]
    \small
    \centering
    \caption{Combined results of spatial accuracy and CLIP score on RLHF settings. 
    }
    \label{tab:RLHF_results}
    \resizebox{.95\linewidth}{!}{%
    \begin{tabular}{@{}llcc@{}}
        \toprule
        Score Type & Method & SDv2 & SDv2(LoRA) \\
        \midrule
        \multirow{3}{*}{Spatial Accuracy(\%)} 
        & Direct & 18.75 & 18.75 \\
        & RLHF & 22.25 & 24.00  \\
        & \textbf{IPR-RLHF} & \textbf{27.00} & \textbf{26.00} \\
        \midrule
        \multirow{3}{*}{CLIP Score} 
        & Direct & 25.75 & 25.75 \\
        & RLHF & 26.00 & 25.72 \\
        & \textbf{IPR-RLHF} & \textbf{26.06} & \textbf{26.10} \\
        \bottomrule
    \end{tabular}
    }
\end{table} 

\section{Experiments}
In this section, we assess IPR's effectiveness through multiple settings. We first introduce our experiment settings. Then, we present the quantitative results and qualitative results of IRP. Finally, we present ablation studies to fully analyze our method.
\subsection{Experimental Settings}
\label{sec:settings and baselines}
\myparagraph{Settings and baselines} Our setting is mainly based on SDv2 \citep{rombach2022highresolution} and SDXL \citep{podell2023sdxl}. Specifically, we use:
(1,2) Two variants of standard SDv2 fine-tuning.
(3) SDv2 LoRA fine-tuning, by freezing the model but training the rank-decomposition matrices injected into the UNet and text-encoder \citep{radford2021learning}.
(4) SDXL LoRA fine-tuning.
In each of the settings, we train 3 epochs for each iteration. Other experimental details are provided in \Cref{sec:suppl_experimental_details}.
We use GLIPv2 \citep{zhang2022glipv2} as the detection model to assign a rescaling reward for the text-image pair. We refer to this as Reinforcement Learning with Detection Feedback (RLDF) in our later experiments.

We compare with the existing RL-trained diffusion baseline: Reinforcement Learning with Human Feedback (RLHF).
We use ImageReward \citep{xu2023imagereward} as the human feedback reward model. This model was pre-trained on large paired datasets that have been annotated by humans. We use the Reward Feedback Learning (ReFL) method for fine-tuning diffusion models.

\myparagraph{Dataset} We use 100 self-training prompts from the VISOR benchmark  \citep{gokhale2023benchmarking}, a challenging dataset focusing on spatial relations. 

\myparagraph{Metrics} We use GLIPv2 for evaluation. This model detects objects and represents them as (object name, bounding box) pairs. It then evaluates accuracy by verifying both the correct count of detected objects and their spatial relationships, determined by comparing the bounding box centers. Images are classified as correct if both object count and spatial relations are accurate; otherwise, they are deemed incorrect.
Besides the spatial accuracy, we also use the standard evaluation metric CLIP score  \citep{radford2021learning}. 



\begin{table}[t]
    \small
    \centering
        \caption{Ablation Study on the effect of three parts of our method (metric: spatial accuracy). (1) PR: Only applying prompt relabeling on diffusion models. (2) RLDF: Only applying RLDF on diffusion models. (3) PR-RLDF: Fine-tuning diffusion models with RLDF and prompt relabeling for one iteration. (4) IPR-RLDF: Iterative training. IPR-RLDF consistently achieves higher spatial accuracy in different settings.}
    \label{tab:ablation_study}
    \resizebox{.95\linewidth}{!}{%
    \begin{tabular}{@{}llcccc@{}}
        \toprule
        Score Type & Method  & SDv2(1) & \makecell{SDv2\\(LoRA)} & \makecell{SDXL\\(LoRA)} & SDv2(2)\\
        \midrule
        \multirow{3}{*}{\makecell{Spatial\\Accuracy (\%)}} 
        & PR & 24.50 & 21.50 & 28.25 & - \\
        & RLDF & 21.50 & 22.00 & 29.75 & 22.44\\
        & PR-RLDF & 25.75 & 24.25 &30.00 & 25.22\\
        & \textbf{IPR-RLDF} & \textbf{28.50}& \textbf{25.25} & \textbf{31.25} & \textbf{32.22} \\
        \bottomrule
    \end{tabular}
    }
\end{table}


\begin{table}[t]
    \small
    \centering
    \caption{Different kinds of spatial accuracy (left-right, above-below, and object number) of our model and baselines in RLDF settings, show that IPR-RLDF outperforms other baselines in recognizing a variety of spatial positional relationships, even though there are minor fluctuations in the above-below spatial relationship of fine-tuned SDXL model.}
    \label{tab:different accuracy types}
    \resizebox{.9\linewidth}{!}{%
    \begin{tabular}{@{}llcccc@{}}
        \toprule
        Spatial Accuracy (\%)  & Method & SDv2(1) & \makecell{SDv2\\(LoRA)} & \makecell{SDXL\\(LoRA)} \\
        \midrule
        \multirow{3}{*}{Left-right } 
        & Direct & 20.63 & 20.63 & 25.00 \\
        & PR & 20.63 &16.88 & 30.63 \\
        & RLDF & 18.13 &18.75 & 28.13 \\
        & PR-RLDF & 20.00 & 22.50 & 29.38 \\
        & \textbf{IPR-RLDF} & \textbf{26.25}& \textbf{23.13} & \textbf{36.25}\\
        \midrule
        \multirow{3}{*}{Above-below } 
        & Direct & 17.50 & 17.50 & 28.33 \\
        & PR & 27.08 & 24.58 & 26.67 \\
        & RLDF & 23.75 & 24.17 & \textbf{30.83} \\
        & PR-RLDF & 29.58 & 25.42 & 30.42 \\
        & \textbf{IPR-RLDF} &\textbf{30.00} & \textbf{26.67} & 27.92 \\
        \midrule
        \multirow{3}{*}{Object number} 
        & Direct & 46.75 & 46.75 & 51.00 \\
        & PR & 53.00 & 48.75 & 54.75 \\
        & RLDF & 46.75 & 48.00 & 56.50 \\
        & PR-RLDF & 52.75 & 47.75 & 51.50 \\
        & \textbf{IPR-RLDF} &\textbf{58.75} & \textbf{49.75} & \textbf{58.25}\\
        \bottomrule
    \end{tabular}
    }
\end{table}

\subsection{Quantitative Results}

\myparagraph{Spatial accuracy} The result shows that IPR-RLDF significantly outperforms the baseline across all settings and suggests that IPR is an effective algorithm in training diffusion models to accurately depict spatial relationships, outperforming conventional training without language feedback or iterative processes. The consistent improvement of overall settings also verifies the method's robustness in different model training and techniques.

\myparagraph{Comparison between IPR-RLDF and IPR-RLHF} \Cref{tab:RLHF_results} demonstrates that IPR-RLHF outperforms RLHF method in both spatial accuracy and CLIP score across two settings, illustrating the effectiveness of the iterative prompt relabeling method with diverse feedback. Although IPR-RLDF achieves a similar increase compared to the IPR-RLHF method (\Cref{tab:combined_results,tab:RLHF_results}), it can get feedback easily from a general detection model, rather than a human feedback reward model specifically trained for the text-image alignment task. 


\subsection{Qualitative Results}
\begin{figure*}[ht]
    \centering
    \includegraphics[width=0.95\linewidth]{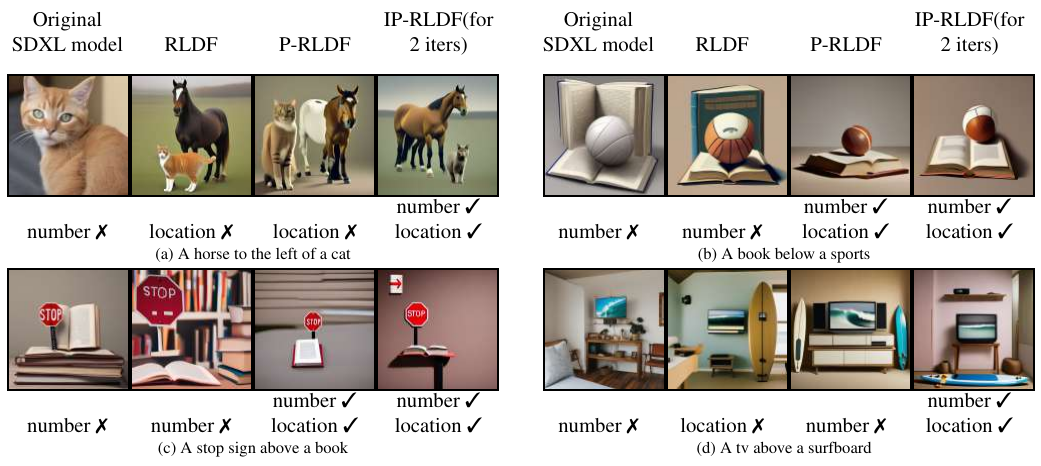}
  \caption{Visual comparison of the original SDXL model with fine-tuned versions using RLDF, PR-RLDF, and IPR-RLDF, across four different prompts. Our algorithm demonstrates superior spatial awareness and accuracy in object depiction, while sacrificing some details.
  }
  \label{fig:visualization}
\end{figure*}
\Cref{fig:visualization} provides a visual comparison of the original SDXL model with fine-tuned versions using RLDF, PR-RLDF, and IPR-RLDF, across four different prompts.
The original SDXL model frequently misinterprets the number and placement of objects, a challenge also observed in RLDF. In contrast, it is evident that our algorithm outperforms both RLDF and the original SDXL model in these aspects, showcasing enhanced spatial awareness and accuracy in depicting the specified objects, while sacrificing some details.

\begin{table}[t]
    \small
    \centering
    \caption{Spatial accuracy of models in RLHF settings, showing that our method outperforms other baselines in recognizing a variety of spatial positional relationships.}
    \label{tab:different accuracy types of RLHF}
    \resizebox{.8\linewidth}{!}{%
    \begin{tabular}{@{}llcc@{}}
        \toprule
        Spatial Accuracy (\%)  & Method & SDv2 & SDv2 (LoRA) \\
        \midrule
        \multirow{3}{*}{Left-right } 
        & Direct & 20.63 & 20.63 \\
        & RLHF & \textbf{21.88} & 18.13 \\
        & \textbf{IPR-RLHF} & \textbf{21.88}& \textbf{21.25} \\
        \midrule
        \multirow{3}{*}{Above-below } 
        & Direct & 17.50 & 17.50 \\
        & RLHF & 22.50 & 27.92 \\
        & \textbf{IPR-RLHF} & \textbf{30.42}& \textbf{29.17} \\
        \midrule
        \multirow{3}{*}{Object number} 
        & Direct & 46.75 & 46.75 \\
        & RLHF & 47.75 & 48.75 \\
        & \textbf{IPR-RLHF} & \textbf{52.25}& \textbf{51.25} \\
        \bottomrule
    \end{tabular}
    }
\end{table}
\vspace{1mm}
\begin{table}[t]
    \centering
    \small
    \caption{Comparison of spatial accuracy at varying GLIP score thresholds for both prompt relabeling and evaluation. Although the performance at a GLIP score threshold of 0.6 is not as good as at 0.4, our algorithm consistently improves spatial accuracy across different detection thresholds.}
    \label{tab:glip score}
    \begin{tabular}{@{}c|ccc@{}}
        \toprule
        \multirow{2}{*}{Thresholds } & \multicolumn{3}{c}{Spatial Accuracy (\%)} \\
        & Direct & PR-RLDF & IPR-RLDF \\
        \midrule
        0.45 & 18.75 & 25.75 & \textbf{28.50} \\
        0.60 & 16.50 & 20.00 & \textbf{20.75} \\
        \bottomrule
    \end{tabular}
\end{table}
    


\subsection{Ablation Study}
\myparagraph{Method breakdown} In \Cref{tab:ablation_study}, we conduct an ablation study to assess the impact of three components of our method, focusing on SDv2 and SDXL settings. The results clearly show that each component is essential to our algorithm's performance. Both prompt relabeling and iterative training significantly improve RLDF outcomes.

\myparagraph{Spatial relationship type study} To further investigate how our algorithm performs across a variety of spatial relationships, we apply an ablation study to different spatial relationships across RLDF and RLHF settings, with a focus on the SDv2 and SDXL models. 
The findings, detailed in \Cref{tab:different accuracy types,tab:different accuracy types of RLHF}, reveal that our model consistently outperforms the baseline across diverse spatial relations. 



\myparagraph{GLIP score thresholds} 
To investigate the impact of the detection model on the final results, we conduct experiments on the SDv2 model, using different thresholds for the GLIP score during detection. For prompt relabeling, we change the thresholds of the glip score from 0.45 to 0.6. We present the results in \cref{tab:glip score}.
We found that when the threshold for the GLIP score is set to 0.6, the performance is not as good as 0.45. This is likely because when GLIPv2 only detects objects with a GLIP score above 0.6, the alignment between the image and the relabeled prompt decreases. Nevertheless, our algorithm consistently improves the spacial accuracy over different detection thresholds.

Additional ablation studies on reward assignments, unseen prompts, color tasks
and rephrased prompts can be found in \Cref{sec:additional_ablation}.

\section{Conclusion}
In this work, we present IPR in response to the challenging task of spatial location generation. IPR is a novel algorithm that designs rich language feedback and incorporates it with the detection model rewards. Then it trains the diffusion model by iteratively receiving rewards and language feedback. This algorithm is a plug-and-play method that is applicable to a range of diffusion models. Extensive results show the model's effectiveness on the challenging spatial relationship benchmark across various settings.
\section*{Limitations}
Our results demonstrate impressive out-of-distribution generation abilities. However, we haven't fully studied the mechanism of how self-training induces such ability. We leave this for future experiments.

\bibliography{anthology,custom}
\bibliographystyle{acl_natbib}

\clearpage
\appendix

\section{Supplementary Experimental Details}\label{sec:suppl_experimental_details}

\subsection{Method Details}\label{sec:additional_method_details}
\begin{figure*}[ht]
    \centering
    \includegraphics[width=\linewidth]{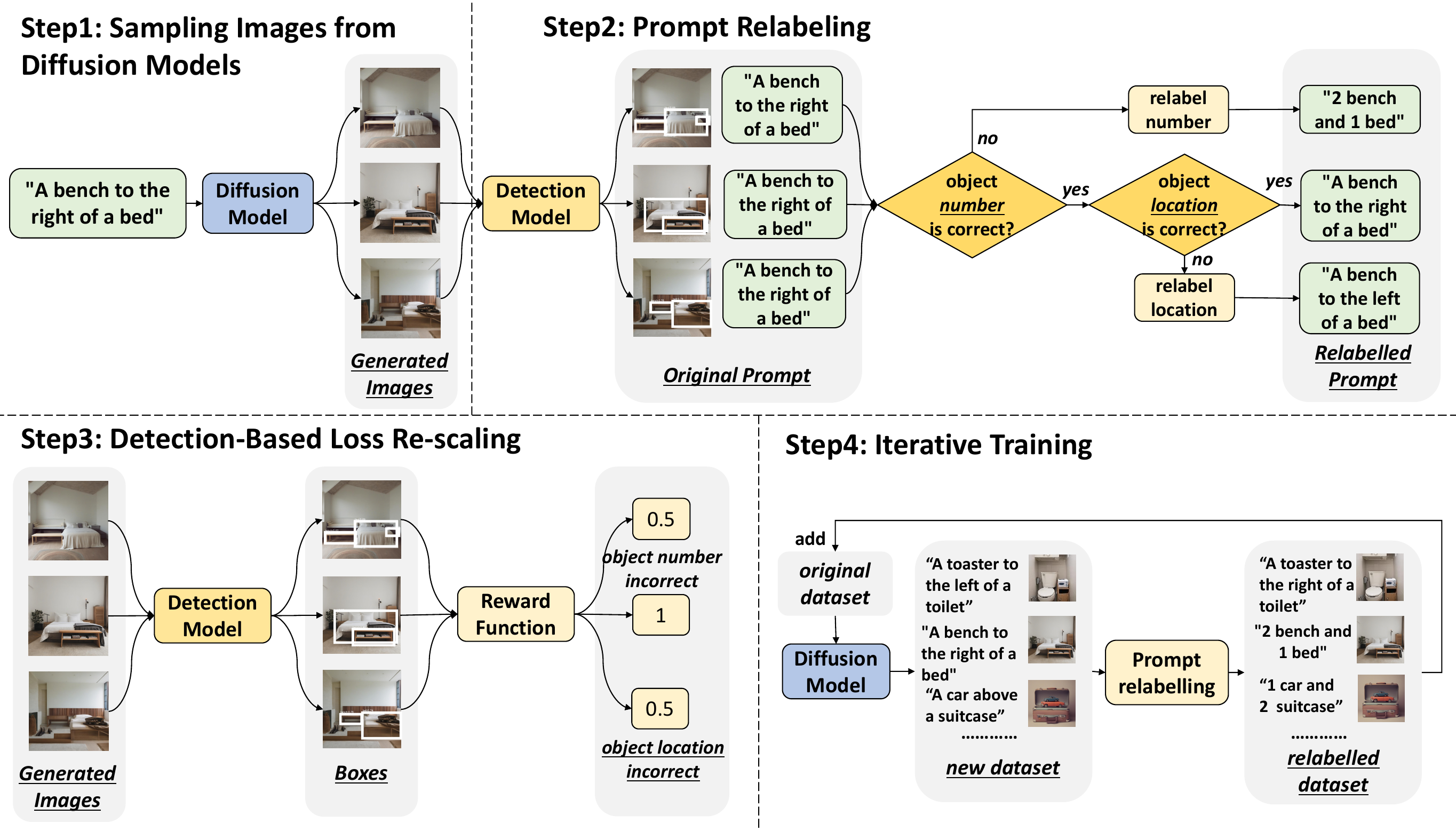}
    \caption{The process of our IPR algorithm. (1) Sampling Images from Diffusion Models: sample images from a diffusion model conditioned on textual prompts. (2) Prompt Relabeling: detect the generated image to yield a bounding box; analyze the box to modify original prompts.
    (3) Detection-Based Loss Re-scaling: apply a detection model to rescale the loss function. (4) Iterative Training: retrain the model with the updated dataset iteratively.
}
    \label{fig:components of pipeline}
\end{figure*}

We provide a breakdown of our method \Cref{fig:components of pipeline} and pseudocode in \Cref{alg:ipRLDF} to better illustrate our method.

\begin{algorithm}[ht]
    \caption{Iterative Prompt Relabeling (IPR)}
    \label{alg:ipRLDF}
    \begin{algorithmic}[1]
    \small
    \Require Text prompts $C$, pre-trained model $\theta$, detection model $D$, iterations $T$, data $X$
    \Ensure Refined parameters $\theta'$
    
    \State Initialize $\theta' \gets \theta$
    
    \For{$\text{iteration} = 1$ \textbf{to} $T$}
        \State $X_0 \gets \text{Sample\_Images}(\text{Diffusion\_Model}, C, \theta')$
        \For{each $x_0$ in $X_0$}
            \State $\text{Objects}, \text{Bounding\_Box} \gets D(x_0)$
            \State $\text{Count}, \text{Relation} \gets \text{Analyze}(D)$
            \State $\text{Reward} \gets \text{CalcReward}(\text{Count}, \text{Relation}, C)$
            \State $\lambda \gets (\text{Count}, \text{Relation} \text{ match } C) ? 1 : 0.5$
            \State $L_{\text{rescaled}} \gets L_{\text{DDPM}}(x_0, \theta') \times \lambda$
            \State $c_{\text{new}} \gets \text{Relabel}(C, D)$
            \State Update $X$ with $(x_0, c_{\text{new}}, L_{\text{rescaled}})$
        \EndFor
        \State $\theta' \gets \text{TrainModel}(X, \theta')$
    \EndFor
    
    \Return $\theta'$
    \end{algorithmic}
\end{algorithm}

\subsection{Additional Ablation Studies}\label{sec:additional_ablation}

\myparagraph{Reward assignment} To examine the impact of rescaling factor $\lambda$, we conduct ablation experiments on the SDv2 model (fine-tuned with LoRA), using different values of $\lambda$ (0.1, 0.5, and 0.7). In \Cref{tab:lambda ablation}, we find that 0.5 yields the best performance, yet, the model is not very sensitive towards different $\lambda$ values.

\myparagraph{Unseen prompts} We assess the performance of our fine-tuned models on prompts that are unseen during self-training, we randomly select 100 new prompts from the VISOR benchmark different from the 100 in the self-training process and test on our trained LoRA fine-tuned SDv2 model. As shown in \Cref{tab:unseen prompts}, our model can also outperform the original pre-trained model in spatial accuracy on the unseen prompts. This demonstrates that such training can be generalized to unseen prompts. 

\begin{figure*}[htbp]
    \centering
    \includegraphics[width=0.9\linewidth]{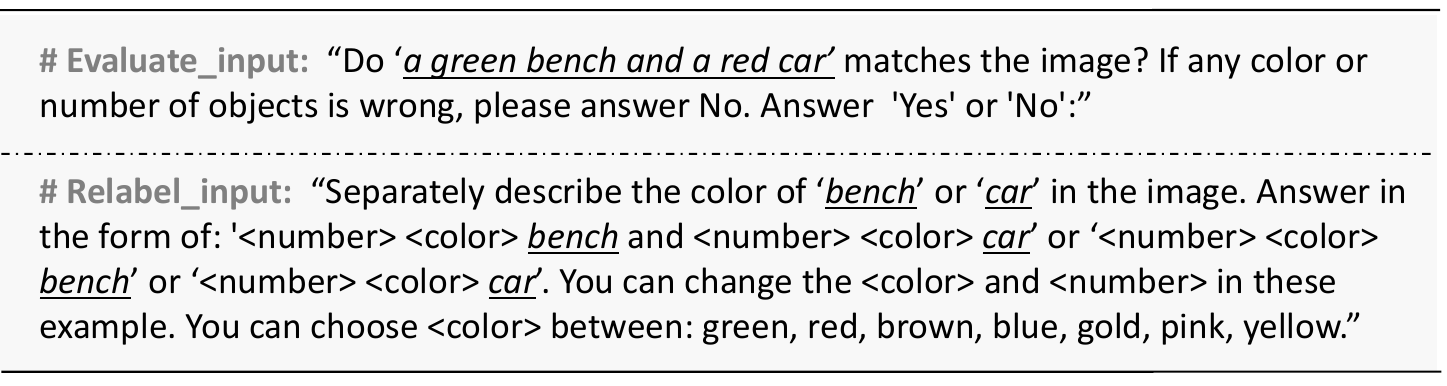}
    \caption{An example prompt for evaluation and prompt relabeling. The evaluating input is directly passed to LLaVA. If the answer is ``No'', we will pass the relabeling input to LLaVA.}
    \label{fig:llava_input}
\end{figure*}

\myparagraph{Color accuracy} In this work, we primarily concentrate on spatial location generation. We study whether IRP can be applicable to other tasks as well. Therefore, we provide some experiments on the color category of T2I-CompBench \citep{huang2023t2icompbench}, a comprehensive benchmark designed for open-world compositional text-to-image generation tasks. Instead of a detection model, we use the LLaVA \citep{liu2023visual} model,  a large-scale multimodal model that combines a vision encoder with an LLM for visual and language understanding, to relabel and evaluate. \Cref{fig:llava_input} shows an example of input prompts used for evaluation and prompt relabeling. The results presented in \Cref{tab:color} demonstrate that our relabeling and iterative training method can generalize to new tasks like color accuracy.


\myparagraph{Generalization abilities on other text inputs}
To evaluate the generalization capabilities of our approach across diverse text inputs, we use GPT-3.5-turbo to rephrase the 100 prompts into other formats. (e.g. “a car above a suitcase” is changed to “positioned above a suitcase is a car"; "an airplane to the right of a clock" is changed to "a clock with an airplane on its right side") Then, we evaluate our fine-tuned SDv2 models on the 100 rephrased prompts.  \cref{tab:rephrased_eval} shows that our method can be generalized to other formats of text inputs as well.

\myparagraph{Training on rephrased prompts}
We also conduct experiments on finetuning SDv2 with the 100 rephrased prompts. At each iteration, we relabel the prompts using our original method. Then we use GPT-3.5-turbo to rephrase the relabeled prompts into other formats. Then, we finetune the model using these rephrased prompts. After the model has been trained, we evaluate the finetuned models using the original 100 prompts. In \cref{tab:rephrased_training}, we find that training on rephrased prompts can also enhance the model's spatial accuracy on original prompts. 

\begin{figure*}[htbp]
    \centering
    \includegraphics[width=1\linewidth]{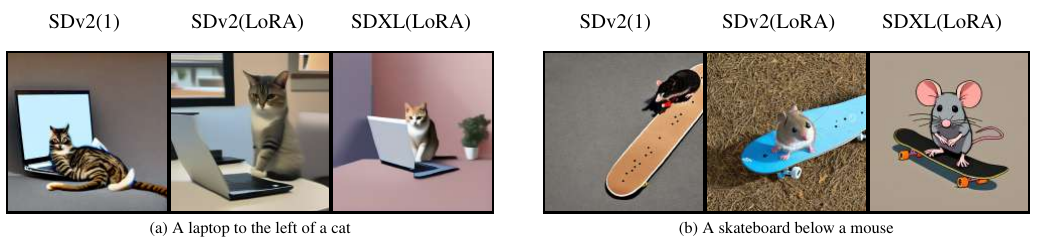}
  \caption{Samples from different models fine-tuned with IPR-RLDF, generated by two distinct prompts. (1) Left column: unfrozen fine-tuning on SDv2. (2)Mid column: using LoRA to fine-tune SDv2. (3) Right column: using LoRA to fine-tune SDXL. The LoRA training exhibits more notable image fidelity than unfrozen training.}
  \label{fig:model_visualization}
\end{figure*}

\begin{table}[t]
    \small
    \centering
    \caption{Ablation study of $\lambda$ on SDv2 model (fine-tuned with LoRA).}
    \label{tab:lambda ablation}
    \begin{tabular}{@{}llccc@{}}
        \toprule
        Score Type & $\lambda$ & Direct & RLDF & PR-RLDF\\
        \midrule
        \multirow{3}{*}{Spatial Acc(\%)} & 0.1 & 18.75 & \textbf{21.75} & 20.00 \\
        & 0.5 & 18.75 & 22.00 & \textbf{24.25} \\
        & 0.7 & 18.75 & 21.25 & \textbf{24.00} \\
        \bottomrule
    \end{tabular}
\end{table}
    
\begin{table}[t]
    \small
    \centering
    \caption{Spatial accuracy of unseen prompts on SDv2 model (fine-tuned with LoRA).}
    \label{tab:unseen prompts}
    \begin{tabular}{@{}lccc@{}}
        \toprule
        Score Type & Direct & \makecell{IPR-RLDF\\(iter2)} & \makecell{IPR-RLDF\\(iter3)}\\
        \midrule
        Spatial Acc(\%) & 17.00 & \textbf{19.25} & 18.75 \\
        \bottomrule
    \end{tabular}
\end{table}

\begin{table}[t]
    \centering
    \small
    \caption{Application of IPR on color accuracy, using LLaVa model to relabel and evaluate, fine-tuning SDv2 with LoRA.}
    \label{tab:color}
    \resizebox{\linewidth}{!}{%
    \begin{tabular}{@{}lccc@{}}
        \toprule
        Score Type & Direct & IPR-RLDF (iter2)\\
        \midrule
        \makecell{LLaVA Evaluation Acc(\%)} & 64.00 & \textbf{66.50} & \\
        \bottomrule
    \end{tabular}
    }
\end{table}


\begin{table}[t]
    \small
    \centering
    \caption{The evaluation results of our approach on rephrased prompts.}
    \label{tab:rephrased_eval}
    \begin{tabular}{@{}lcc@{}}
        \toprule
        Score Type & Direct & PR-RLDF \\
        \midrule
        Spatial Acc (\%) & 16.25 & \textbf{17.50} \\
        \bottomrule
    \end{tabular}
\end{table}

\begin{table}[t]
    \small
    \centering
    \caption{Comparison between training on original and rephrased Prompts.}
    \label{tab:rephrased_training}
    \resizebox{\linewidth}{!}{%
    \begin{tabular}{@{}llccc@{}}
        \toprule
        Score Type & Training Prompts & Direct & PR-RLDF & IPR-RLDF \\
        \midrule
        \multirow{2}{*}{Spatial Acc (\%)} & Original Prompts & 18.75 & 25.75 & \textbf{28.50} \\
        & Rephrased Prompts & 18.75 & 21.50 & \textbf{23.00} \\
        \bottomrule
    \end{tabular}
    }
\end{table}

\subsection{Additional Quantitative Results}\label{sec:additional_quantitative}

\begin{figure*}[htbp]
    \centering
    \includegraphics[width=1\linewidth]{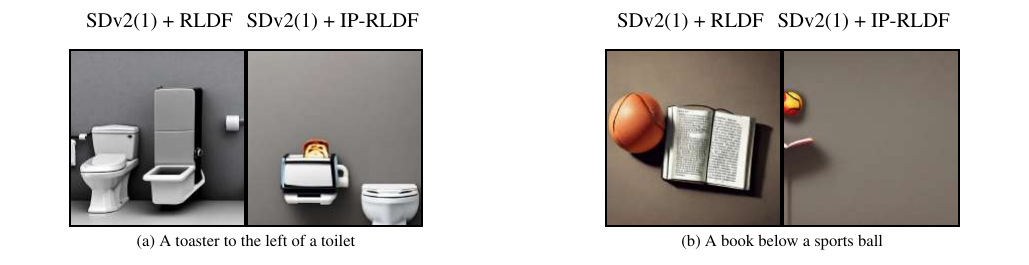}
  \caption{Samples from different fine-tuned models. (1) Left column: unfrozen fine-tuning on SDv2 with RLDF. (2) Right column: unfrozen fine-tuning on SDv2 with IPR-RLDF.  Compared with RLDF, the SDv2 model fine-tuned with IPR-RLDF exhibits notable image fidelity but lower spatial accuracy.}
\label{fig:sdv2_fidelity_visualization}
\end{figure*}

\myparagraph{Text-to-image alignment} 
\label{sec:Text-to-Image Alignment Analysis} IPR-RLDF also demonstrates remarkable gains in CLIP score over both the RLDF algorithm and the original text-to-image models in SDv2 (LoRA) and SDXL (LoRA) settings, indicating that it keeps the image's quality and achieves remarkable text-to-image alignment while enhancing the spatial correctness. One exception occurs during SDv2(1), where the RLDF outperforms IPR-RLDF on the CLIP score. This is likely due to a trade-off between spatial accuracy and overall image quality, with IPR-RLDF achieving higher spatial accuracy at the expense of a slight decrement in CLIP score. As shown in \cref{fig:sdv2_fidelity_visualization}, IPR-RLDF blurs and RLDF kept more details, while IPR-RLDF shows higher spatial accuracy.

\myparagraph{LLaVA Evaluation Results}
Besides GLIP, we add another evaluation method. Since recent VLMs have shown comprehensive visual understanding abilities, we use the most recent VLM--the LLaVA model to evaluate spatial accuracy. Specifically, the model is asked to determine the spatial correctness of each input image. The results are shown in \cref{tab:llava_gpt_eval}. Our method exhibits significant improvement over previous methods using LLaVA to evaluate. 
\begin{table}[t]
    \small
    \centering
    \caption{\textbf{LLaVA and GPT-4-turbo Evaluation Results.} Our method exhibits significant improvement over previous methods using LLaVA and GPT-4-turbo to evaluate. }
    \label{tab:llava_gpt_eval}
    \resizebox{\linewidth}{!}{%
    \begin{tabular}{@{}lccc@{}}
        \toprule
        Metric & Direct & PR-RLDF & IPR-RLDF \\
        \midrule
        LLaVA Evaluation Acc (\%) & 49.50 & 52.00 & \textbf{53.50} \\
        GPT-4-turbo Evaluation Acc (\%) & 9.00 & 13.00 & \textbf{19.00} \\
        \bottomrule
    \end{tabular}
    }
\end{table}

\myparagraph{GPT Models Evaluation Results}
We also use GPT-4-turbo to evaluate the spatial accuracy of the SDv2 model. As we did in the LLaVA evaluation experiments, the model is asked to determine the spatial correctness of each input image. The results are shown in \cref{tab:llava_gpt_eval}. We found that our method also has an improvement in spatial accuracy.

\myparagraph{Training method analysis} \cref{tab:combined_results} shows a performance difference between different training techniques. In the LoRA setting, where the weight of pre-trained models is frozen and only the rank-decomposition matrices injected are trained, the performance gain is less compared with fine-tuning all parameters. However, the LoRA training exhibits notable image fidelity compared with fine-tuning all parameters, as shown in \cref{fig:model_visualization}. 
This suggests a trade-off between enhancing spatial accuracy and keeping overall image quality, indicating possible directions for improvement in future research.

\subsection{Model Details}

\subsubsection{Stable Diffusion v2 (SDv2)}
Similar to Imagen \citep{ho2022imagen}, Stable Diffusion \citep{rombach2022highresolution} is a latent text-to-image model, with a frozen CLIP ViT-L/14 text encoder \citep{radford2021learning} and an 860M UNet \citep{ronneberger2015unet} constructure. This model was pretrained on 256*256 images followed by fine-tuning on 512*512 images sourced from the LAION 5B dataset \citep{schuhmann2022laion5b}. It excelled in text-to-image tasks, while concurrently supporting image-to-image tasks. Stable Diffusion 2.0, however, employs OpenCLIP-ViT/H as its text encoder, which is trained from scratch. In our experiments, we use Stabilityai/stable-diffusion-2-1-base 
 (512*512 resolution) and Stabilityai/stable-diffusion-2-1 (768*768 resolution) as base models, both fine-tuned on Stable Diffusion 2.0.

\subsubsection{Stable Diffusion XL (SDXL)}
Compared to the previous stable diffusion models, Stable Diffusion XL \citep{podell2023sdxl} features a UNet that is three times larger and integrates OpenCLIP ViT-bigG/14 with the original text encoder. It also introduces crop-conditioning and a two-stage model process to significantly enhance the quality of generated images. Stable Diffusion XL demonstrates improved support for shorter prompts and glyphs in images.

\begin{table*}[t]
    \small
    \centering
    \caption{Comparison between different settings of RLDF and RLHF, from five different aspects: (1) pretrained models. (2) Unfrozen parts of each model during fine-tuning. (3) Resolution of models. (4) The initial learning rate of iterative training. (5) Spacial accuracy improvement of different settings.}
    \label{tab:comparison between settings}
    \resizebox{\linewidth}{!}{%
        \begin{tabular}{@{}llccccc@{}}
            \toprule
            & Settings & Pretrained models & Unfrozen parts & Resolution & Initial lr & Spatial acc(\%)\\
            \midrule
            \multirow{4}{*}{RLDF} 
            & SDv2(1) & Stabilityai/stable-diffusion-2-1 & Full weights & 768 & 1e-6 & 28.50/18.75\\
            & SDv2(LoRA) & Stabilityai/stable-diffusion-2-1 &  Low-Rank Adaptation & 768 & 1e-6 & 25.25/18.75\\
            & SDXL(LoRA) & Stabilityai/stable-diffusion-xl-base-1.0 & Low-Rank Adaptation & 1024 & 1e-5 &31.25/27.00\\
            & SDv2(2) & Stabilityai/stable-diffusion-2-1-base & Full weights & 512 & 1e-5 & \textbf{32.22/17.00}\\
            \midrule
            \multirow{2}{*}{RLHF} &
            SDv2 & Stabilityai/stable-diffusion-2-1 & Full weights & 768 & 1e-6 & 27.00/18.75\\
            & SDv2(LoRA) & Stabilityai/stable-diffusion-2-1 &  Low-Rank Adaptation & 768 & 1e-5 & 26.00/18.75\\
            \bottomrule
        \end{tabular}
    }
\end{table*}

\subsection{Training Details}
\subsubsection{Low-Rank Adaptation (LoRA)} Due to the considerable time and computational resources required for full fine-tuning of large models, Low-Rank Adaptation \citep{hu2021lora} was introduced for fine-tuning in particular tasks. It involves injecting rank decomposition matrices into transformer layers while freezing all other model weights. LoRA attains training quality comparable to full fine-tuning but in less time and with fewer computational resources. Initially deployed in large language models, LoRA can also be extended to cross-attention layers in text-to-image models such as Stable Diffusion, where it has demonstrated outstanding. In our experiments, we use LoRA to fine-tune SDv2 and SDXL, validating that our IPR algorithm serves as an additional plug-and-play algorithm capable of integration across various diffusion model settings.

\myparagraph{Dataset} We use 100 self-training prompts from the VISOR benchmark  \citep{gokhale2023benchmarking}, a challenging dataset focusing on spatial relations. 
In each iteration, we sample 400 data in total to do the training.  If overfitting, only the base dataset will increase by 2000 prompts in each iteration.

\subsubsection{Training Cost} 
In our IPR-RLDF method, when fine-tuning SDv2 with LoRA on four NVIDIA Tesla V100 GPUs, the training time for each iteration is approximately 15 minutes. While in the IPR-RLHF baseline, fine-tuning SDv2 with LoRA on four V100s costs around 60 minutes for each iteration.

\begin{figure*}[ht]
    \centering
    \includegraphics[width=1\linewidth]{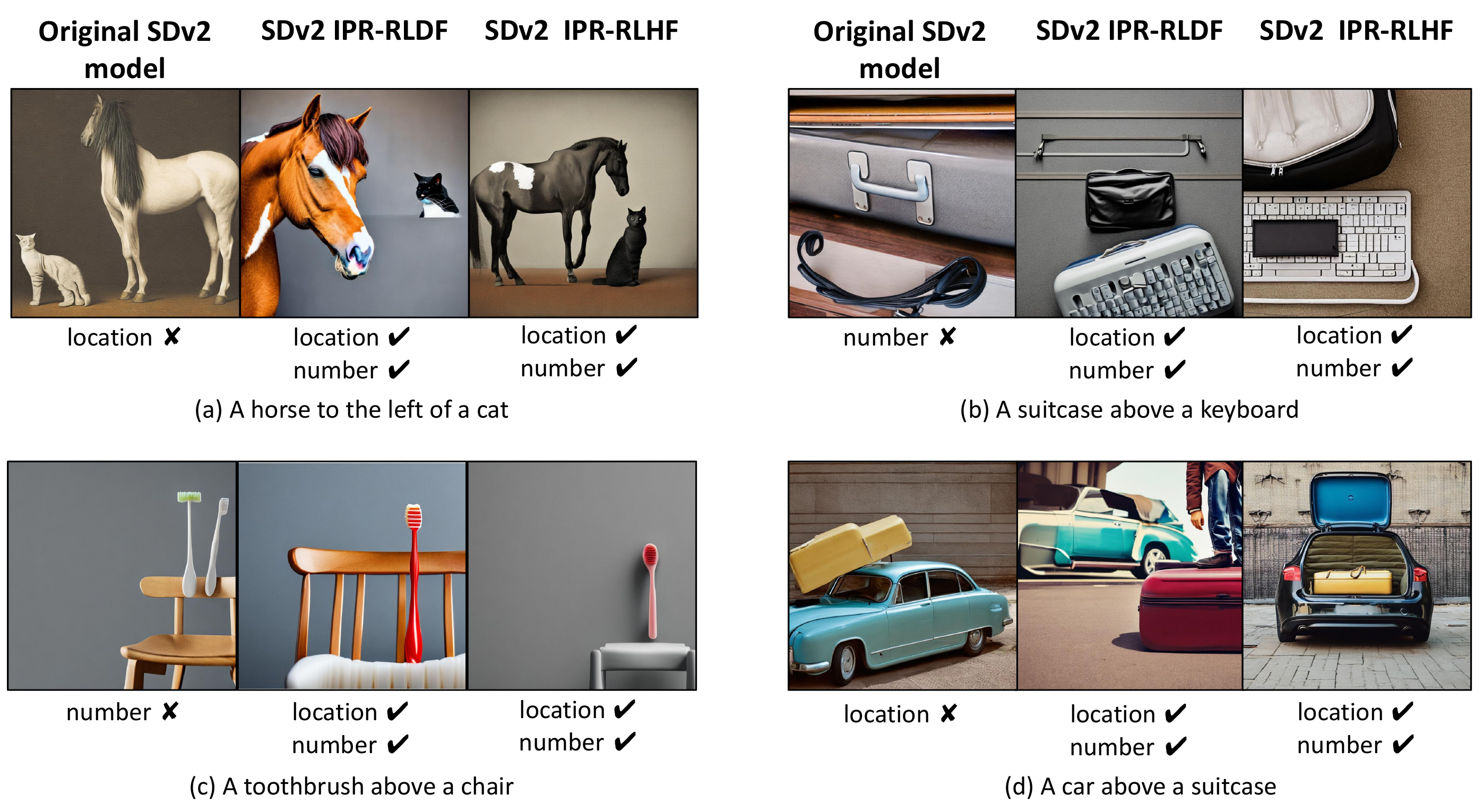}
    \caption{Qualitative examples of different models, generated by four distinct prompts. Models are as follows: (1) Using SDv2 as the original model. (2) IPR-RLDF (LoRA fine-tuning). (3) IPR-RLHF (LoRA fine-tuning). Both IPR-RLDF and IPR-RLHF exhibit commendable spatial accuracy.}
    \label{fig:rlhf_sdv2_lora}
\end{figure*}

\begin{figure*}[ht]
    \centering
    \includegraphics[width=1\linewidth]{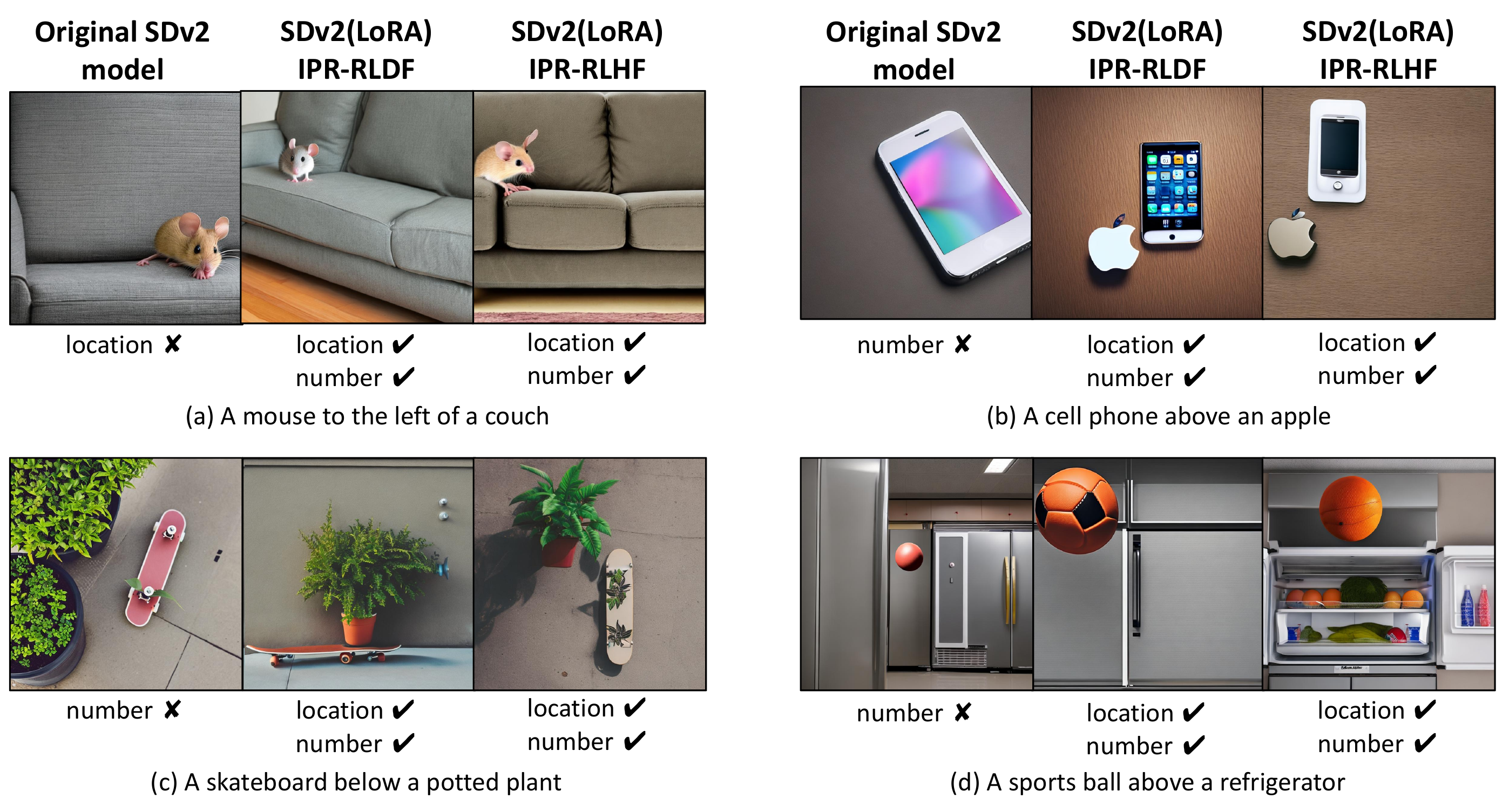}
    \caption{Qualitative examples of different models, generated by four distinct prompts. Models are as follows: (1) Using SDv2 as the original model. (2) IPR-RLDF (unfrozen fine-tuning). (3) IPR-RLHF (unfrozen fine-tuning). Both IPR-RLDF and IPR-RLHF exhibit commendable spatial accuracy.}
    \label{fig:rlhf_sdv2}
\end{figure*}

\subsection{Difference between Fine-tuning Settings} 
In different RLDF and RLHF fine-tuning settings, we employ the previously mentioned models and fine-tuning methods. Detailed differences among settings are presented in \Cref{tab:comparison between settings}. 

\subsection{Detection Model Details}
GLIPv2 \citep{zhang2022glipv2}, a grounded Vision-Language (VL) understanding model, integrates localization \citep{lin2015microsoft, caesar2018cocostuff} and VL understanding \citep{chen2015microsoft, agrawal2016vqa, kiros2014unifying} to establish grounded VL understanding. It achieves this by innovatively transforming localization tasks into a concentration of category names within VL understanding \citep{li2022grounded}. This innovative approach effectively resolves the conflicting output format requirements between localization and VL understanding, allowing them to help each other mutually. Consequently, it attains outstanding performance in both localization and understanding tasks. Leveraging its outstanding zero-shot detection capability, we employ the pretrained GLIPv2 model as the detection component in our framework, yielding excellent results.

\subsection{Human Preference Reward Model}
We use ImageReward \citep{xu2023imagereward} as the reward model in the RLHF baseline. It is the first general-purpose text-to-image human preference reward model, trained with 137k expert comparisons. It proficiently
encodes human preferences. To fine-tune diffusion models with ImageReward, we use Reward Feedback Learning (ReFL) \citep{xu2023imagereward}, an algorithm that fine-tunes LDMs directly. It treats the scores from the reward model as human preference losses, which are then back-propagated to a randomly selected step in the denoising process.

\subsection{Additional Visualizations}
\begin{figure*}
    \centering
    \includegraphics[width=1\linewidth]{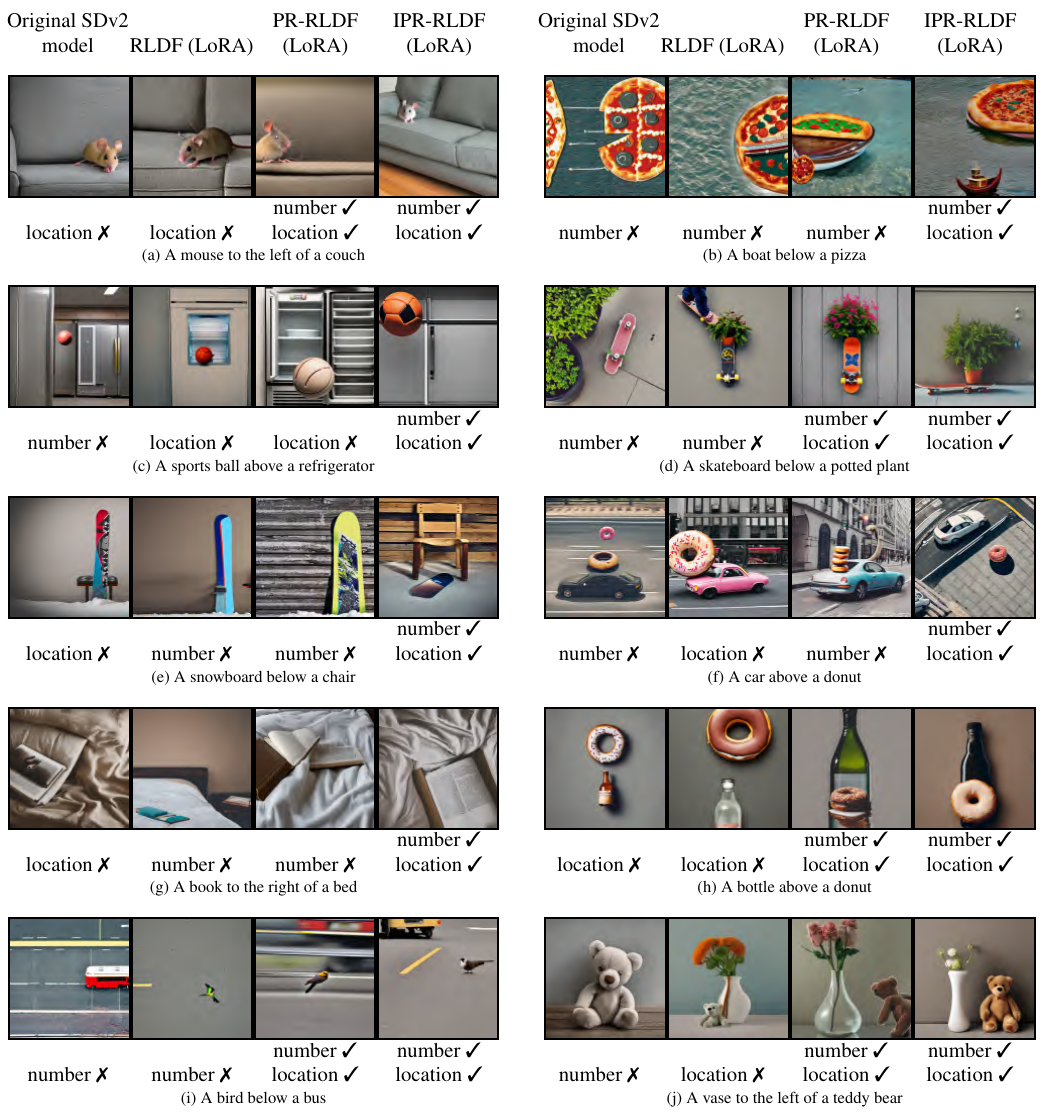}
  \caption{Samples from different models, generated by ten distinct prompts. Models are as follows: (1) Using SDv2 as the original model. (2) RLDF(LoRA fine-tuning). (3) PR-RLDF(LoRA fine-tuning). (4) IPR-RLDF(trained for two iters, LoRA fine-tuning). Compared to the original SDv2 model, RLDF, and PR-RLDF methods, our method showcases enhanced spatial awareness and accuracy in generating images of specified objects.}
  \label{fig:add_visualization_SDv2_lora}
\end{figure*}

\begin{figure*}
    \centering
    \includegraphics[width=1\linewidth]{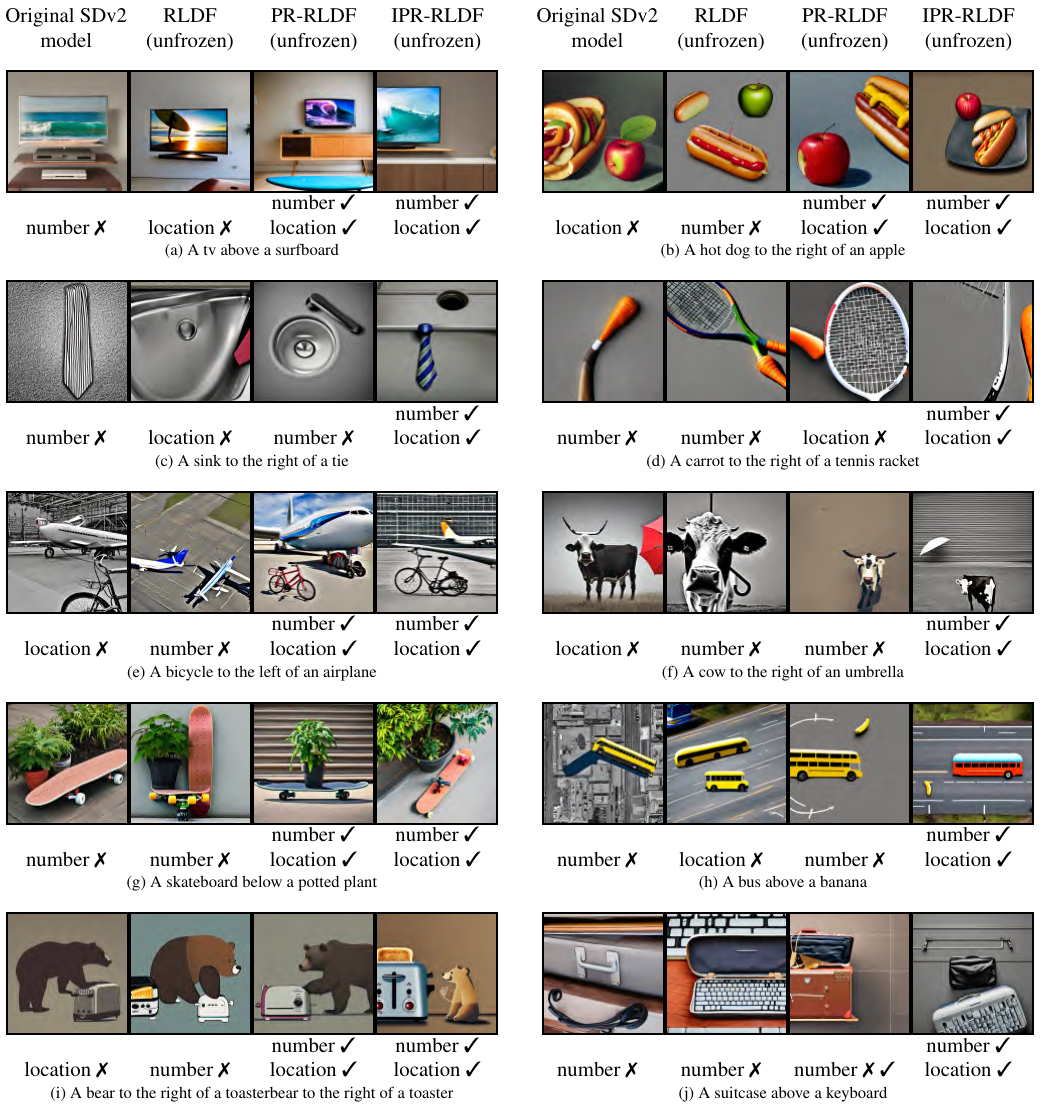}
  \caption{Samples from different models, generated by ten distinct prompts. Models are as follows: (1) Using SDv2 as the original model. (2) RLDF(unfrozen fine-tuning). (3) PR-RLDF(unfrozen fine-tuning). (4) IPR-RLDF(trained for two iters, unfrozen fine-tuning). Compared to the original SDv2 model, RLDF, and PR-RLDF methods, our method showcases enhanced spatial awareness and accuracy in generating images of specified objects.}
  \label{fig:add_visualization_SDv2}
\end{figure*}

\begin{figure*}
    \centering
    \includegraphics[width=1\linewidth]{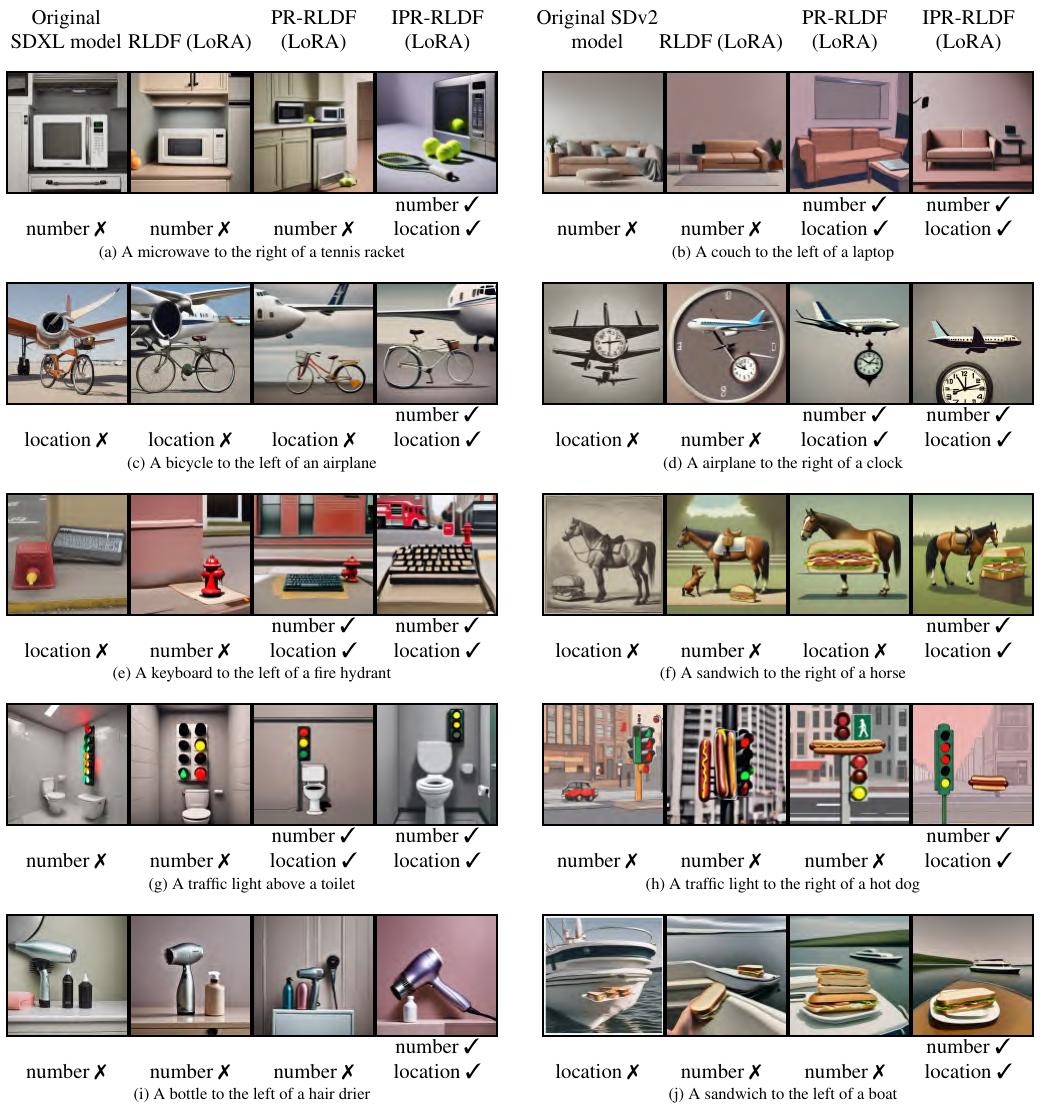}
  \caption{Samples from different models, generated by ten distinct prompts. Models are as follows: (1) Using SDXL as the original model. (2) RLDF(LoRA fine-tuning). (3) PR-RLDF(LoRA fine-tuning). (4) IPR-RLDF(trained for two iters, LoRA fine-tuning). Compared to the original SDXl model, RLDF, and PR-RLDF methods, our method showcases enhanced spatial awareness and accuracy in generating images of specified objects.}
  \label{fig:add_visualization_SDXL}
\end{figure*}

We provide qualitative samples from original models and models fine-tuned with various methods (RLDF, PR-RLDF, and IPR-RLDF). In comparison to the original models, RLDF, and PR-RLDF methods, the IPR-RLDF method notably demonstrates superior spatial accuracy. \Cref{fig:add_visualization_SDv2_lora} shows samples from SDv2 LoRA fine-tuning models; \Cref{fig:add_visualization_SDv2} shows samples from SDv2 unfrozen fine-tuning models; \Cref{fig:add_visualization_SDXL} shows additional samples from SDXL LoRA fine-tuning models. 
Images in each figure are generated by ten distinct prompts. 
Also, there are qualitative examples of IPR-RLDF and IPR-RLDF in \Cref{fig:rlhf_sdv2_lora,fig:rlhf_sdv2}. As is presented, both IPR-RLDF and IPR-RLHF exhibit commendable spatial accuracy.

\section{Additional Related Works}
\subsection{Detection Model}
Many studies have concentrated on object detection tasks \citep{redmon2016you, lin2018focal, chen2019mmdetection, carion2020endtoend}, while the ability to detect certain rare objects \citep{gupta2019lvis} still lacks proficiency. Recent work aims to address this issue, adopting novel approaches such as zero-shot \citep{bucher2019zeroshot}, few-shot \citep{Li_2021_CVPR}, or weakly-supervised \citep{bucher2019zeroshot} methods. Notably, MEDTER \citep{kamath2021mdetr}, GLIP \citep{li2022grounded}, and GLIPv2 \citep{zhang2022glipv2} introduce an innovative perspective by transforming object detection tasks into grounded Vision-Language tasks. This integration of Vision-Language aspects into object detection yields remarkable effects in tasks like few-shot object detection. Thus, in our approach, we employ GLIPv2 as our chosen detection model within the pipeline.



\end{document}